\documentclass{article}
\usepackage{spconf,amsmath,graphicx}
\usepackage{amssymb}
\usepackage{subfig}
\usepackage[hidelinks]{hyperref}
\usepackage{color}
\usepackage{siunitx}
\usepackage{booktabs}
\usepackage{etoolbox}
\usepackage{multirow}
\usepackage{amsmath}
\usepackage{changes}
\newrobustcmd{\B}{\bfseries}
\title{Point Cloud Denoising via Momentum Ascent in Gradient Fields}
\name{ Yaping Zhao$^{1,3,*}$
 , Haitian Zheng$^{2,*}$, Zhongrui Wang$^{1,3}$, Jiebo Luo$^{2}$, Edmund Y. Lam$^{1,3,\dag}$}

\address{
$^1$ The University of Hong Kong, Pokfulam, Hong Kong SAR \, \indent \,
$^2$ University of Rochester, USA\\
$^3$ACCESS –- AI Chip Center for Emerging Smart Systems, Hong Kong SAR}
	
\begin{document}
\newcommand{\Amat}{{\bf A}}
\newcommand{\Bmat}{{\bf B}}
\newcommand{\Cmat}{{\bf C}}
\newcommand{\Dmat}{{\bf D}}
\newcommand{\Emat}[0]{{{\bf E}}}
\newcommand{\Fmat}[0]{{{\bf F}}}
\newcommand{\Gmat}[0]{{{\bf G}}}
\newcommand{\Hmat}[0]{{{\bf H}}}
\newcommand{\Imat}{{\bf I}}
\newcommand{\Jmat}[0]{{{\bf J}}}
\newcommand{\Kmat}[0]{{{\bf K}}}
\newcommand{\Lmat}[0]{{{\bf L}}}
\newcommand{\Mmat}[0]{{{\bf M}}}
\newcommand{\Nmat}[0]{{{\bf N}}}
\newcommand{\Omat}[0]{{{\bf O}}}
\newcommand{\Pmat}[0]{{{\bf P}}}
\newcommand{\Qmat}[0]{{{\bf Q}}}
\newcommand{\Rmat}[0]{{{\bf R}}}
\newcommand{\Smat}[0]{{{\bf S}}}
\newcommand{\Tmat}[0]{{{\bf T}}}
\newcommand{\Umat}{{{\bf U}}}
\newcommand{\Vmat}[0]{{{\bf V}}}
\newcommand{\Wmat}[0]{{{\bf W}}}
\newcommand{\Xmat}{{\bf X}}
\newcommand{\Ymat}[0]{{{\bf Y}}}
\newcommand{\Zmat}{{\bf Z}}

\newcommand{\av}{\boldsymbol{a}}
\newcommand{\Av}{\boldsymbol{A}}
\newcommand{\Cv}{\boldsymbol{C}}
\newcommand{\bv}{\boldsymbol{b}}
\newcommand{\cv}{{\boldsymbol{c}}}
\newcommand{\dv}{\boldsymbol{d}}
\newcommand{\ev}[0]{{\boldsymbol{e}}}
\newcommand{\fv}{\boldsymbol{f}}
\newcommand{\Fv}[0]{{\boldsymbol{F}}}
\newcommand{\gv}[0]{{\boldsymbol{g}}}
\newcommand{\hv}[0]{{\boldsymbol{h}}}
\newcommand{\iv}[0]{{\boldsymbol{i}}}
\newcommand{\jv}[0]{{\boldsymbol{j}}}
\newcommand{\kv}[0]{{\boldsymbol{k}}}
\newcommand{\lv}[0]{{\boldsymbol{l}}}
\newcommand{\mv}[0]{{\boldsymbol{m}}}
\newcommand{\nv}{\boldsymbol{n}}
\newcommand{\ov}[0]{{\boldsymbol{o}}}
\newcommand{\pv}[0]{{\boldsymbol{p}}}
\newcommand{\qv}[0]{{\boldsymbol{q}}}
\newcommand{\rv}[0]{{\boldsymbol{r}}}
\newcommand{\sv}[0]{{\boldsymbol{s}}}
\newcommand{\tv}[0]{{\boldsymbol{t}}}
\newcommand{\uv}[0]{{\boldsymbol{u}}}
\newcommand{\vv}{\boldsymbol{v}}
\newcommand{\wv}{\boldsymbol{w}}
\newcommand{\Wv}{\boldsymbol{W}}
\newcommand{\xv}{\boldsymbol{x}}
\newcommand{\yv}{\boldsymbol{y}}
\newcommand{\Xv}{\boldsymbol{X}}
\newcommand{\Yv}{\boldsymbol{Y}}
\newcommand{\zv}{\boldsymbol{z}}

\newcommand{\Gammamat}[0]{{\boldsymbol{\Gamma}}}
\newcommand{\Deltamat}[0]{{\boldsymbol{\Delta}}}
\newcommand{\Thetamat}{\boldsymbol{\Theta}}
\newcommand{\Lambdamat}{{\boldsymbol{\Lambda}}}
\newcommand{\Ximat}[0]{{\boldsymbol{\Xi}}}
\newcommand{\Pimat}[0]{{\boldsymbol{\Pi}} }
\newcommand{\Sigmamat}{\boldsymbol{\Sigma}}
\newcommand{\Upsilonmat}[0]{{\boldsymbol{\Upsilon}} }
\newcommand{\Phimat}{\boldsymbol{\Phi}}
\newcommand{\Psimat}{\boldsymbol{\Psi}}
\newcommand{\Omegamat}{{\boldsymbol{\Omega}}}

\newcommand{\Lambdav}{\bm{\Lambda}}
\newcommand{\alphav}{\boldsymbol{\alpha}}
\newcommand{\betav}[0]{{\boldsymbol{\beta}} }
\newcommand{\gammav}{{\boldsymbol{\gamma}}}
\newcommand{\deltav}[0]{{\boldsymbol{\delta}} }
\newcommand{\epsilonv}{\boldsymbol{\epsilon}}
\newcommand{\zetav}[0]{{\boldsymbol{\zeta}} }
\newcommand{\etav}[0]{{\boldsymbol{\eta}} }
\newcommand{\thetav}{\boldsymbol{\theta}}
\newcommand{\iotav}[0]{{\boldsymbol{\iota}} }
\newcommand{\kappav}{{\boldsymbol{\kappa}}}
\newcommand{\lambdav}[0]{{\boldsymbol{\lambda}} }
\newcommand{\muv}{\boldsymbol{\mu}}
\newcommand{\nuv}{{\boldsymbol{\nu}}}
\newcommand{\xiv}{{\boldsymbol{\xi}}}
\newcommand{\omicronv}[0]{{\boldsymbol{\omicron}} }
\newcommand{\piv}{\boldsymbol{\pi}}
\newcommand{\rhov}[0]{{\boldsymbol{\rho}} }
\newcommand{\sigmav}[0]{{\boldsymbol{\sigma}} }
\newcommand{\tauv}[0]{{\boldsymbol{\tau}} }
\newcommand{\upsilonv}[0]{{\boldsymbol{\upsilon}} }
\newcommand{\phiv}{\boldsymbol{\phi}}
\newcommand{\chiv}[0]{{\boldsymbol{\chi}} }
\newcommand{\psiv}{\boldsymbol{\psi}}
\newcommand{\omegav}[0]{{\boldsymbol{\omega}} }

\newcommand{\xin}[1]{{\textcolor{red}{#1}}}

\newcommand{\ts}{^{\top}}
\newcommand{\TV}{{\rm TV}}
\newtheorem{definition}{Definition}
\newtheorem{lemma}{Lemma}
\newtheorem{corollary}{Corollary}
\newtheorem{theorem}{Theorem}%[section]

\twocolumn[{%
\renewcommand\twocolumn[1][]{#1}%
\maketitle
\begin{center}
\vspace{-20pt}
    \includegraphics[width=1.\linewidth]{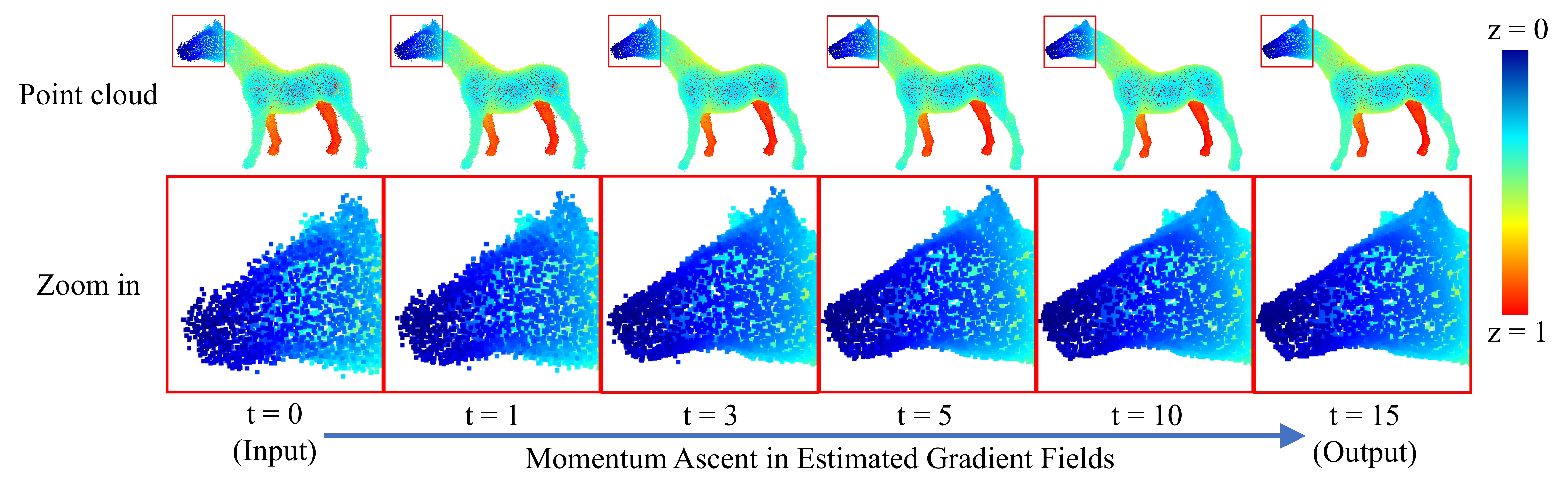}
    \vspace{-15pt}
	\captionof{figure}{
	Given a noisy point cloud as input,
we utilize a neural network to estimate the implicit gradient fields, and iteratively update the position of each point by momentum gradient ascent towards the underlying surface. The color of each 3D point varies according to its $z$ coordinate (perpendicular to the plane). 
% After iterations, the noisy point cloud would finally converge to a clean one.
	}
	\label{fig:teaser}
	\vspace{10pt}
\end{center}
}
]

\newcommand\blfootnote[1]{%
  \begingroup
  \renewcommand\thefootnote{}\footnote{#1}%
  \addtocounter{footnote}{-1}%
  \endgroup
}

\blfootnote{$^{*}$Equal contribution.  $^{\dag}$Corresponding author.}
\blfootnote{This work is supported in part by the Research Grants Council (GRF 17201620), by the Research Postgraduate Student Innovation Award (The University of Hong Kong), and by ACCESS –- AI Chip Center for Emerging Smart Systems, Hong Kong SAR.}

% \maketitle

% \begin{figure*}
%     \includegraphics[width =\linewidth]{img/teaser1.png}
%     \caption{Illustration of the human-centric RefSR that transfers the high-definition human body details onto low resolution video.}
%     \label{fig:teaser}
%     \vspace{-10pt}
% \end{figure*}
\begin{abstract}
To achieve point cloud denoising, traditional methods heavily rely on geometric priors, and most learning-based approaches suffer from outliers and loss of details. Recently, the gradient-based method was proposed to estimate the gradient fields from the noisy point clouds using neural networks, and refine the position of each point according to the estimated gradient. However, the predicted gradient could  fluctuate, leading to perturbed and unstable solutions, as well as a long inference time. To address these issues, we develop the momentum gradient ascent method that leverages the information of previous iterations in determining the trajectories of the points, thus improving the  stability of the solution and reducing the inference time.
Experiments demonstrate that the proposed method outperforms state-of-the-art approaches with a variety of point clouds, noise types, and noise levels.
Code is available at: \href{https://github.com/IndigoPurple/MAG}{\textcolor{blue}{https://github.com/IndigoPurple/MAG}}.

\end{abstract}


\section{introduction}
\label{sec:intro}

Point cloud denoising aims to restore clean point clouds from noise-corrupted ones. Due to the inherent limitations of acquisition devices or matching ambiguities in the 3D reconstruction, noise inevitably degrades the quality of scanned or reconstructed point clouds, for which point cloud denoising is favored. Moreover, the quality of point clouds affect the performance of downstream 3D vision tasks, \textit{e.g.}, detection and segmentation. Therefore, point cloud denoising offers crucial preprocessing for relevant 3D vision applications.

Unlike image and video denoising~\cite{zhao2022manet}, point cloud denoising is challenging because of the intrinsic unordered
characteristic of point clouds. Point clouds consist of discrete 3D points irregularly sampled from continuous surfaces. Perturbed by noise, the 3D points can deviate from their original positions and yield the wrong coordinates. 
To tackle this issue, both traditional ~\cite{alexa2001point,avron2010,cazals2005estimating, huang2013edge,mattei2017point,sun2015denoising} and deep learning~\cite{duan20193d, hermosilla2019total,pistilli2020learning,rakotosaona2020pointcleannet} methods have been explored but show limited performance. Recently, Luo and Hu propose score-based denoising (Score)~\cite{luo2021score} to iteratively update the point positions according to the estimated gradient fields. However, the predicted gradients can fluctuate, leading to perturbed and unstable solutions, as well as a large inference time. 

\begin{figure*}
\centering
\vspace{-10pt}
    \includegraphics[width =0.9\linewidth]{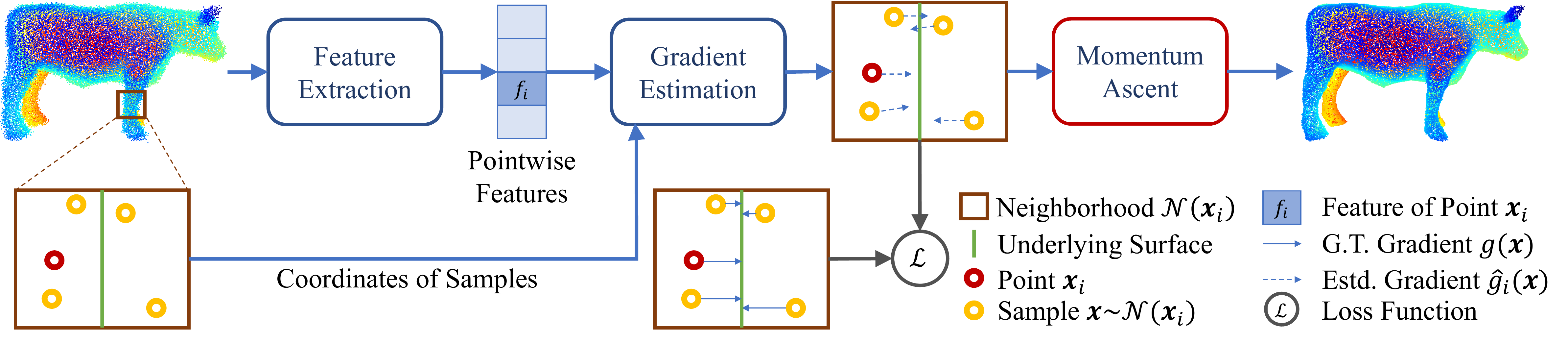}
    \vspace{-5pt}
    \caption{Pipeline of point cloud denoising via momentum ascent in gradient fields. 
    % We first employ Score~\cite{luo2021score} to extract feature and estimate gradients, and then utilize the estimated gradients to denoise point clouds by momentum ascent.
    }
    \label{fig:method}
\end{figure*}

To improve the performance and efficiency of the gradient-based method, as Figure~\ref{fig:teaser} shows, we propose a novel iterative paradigm of point cloud denoising motivated by the classical momentum method~\cite{goodfellow2016deep} in optimization. Specifically, we employ Score~\cite{luo2021score} to estimate the gradient fields from the noisy point clouds, and compute the displacement of each point according to the estimated gradient. To avoid the fluctuated gradient of Score, we apply a momentum gradient ascent method that utilizes the previous iterations to seek promising directions to move forward, thus improving the solution stability and inference time. Experiments demonstrate that the proposed method outperforms state-of-the-art approaches. 

Our main contributions are as follows:
\begin{enumerate}
    \item We propose a \textbf{simple yet effective} iterative paradigm of point cloud denoising, which leverages past gradients to seek promising directions to move forward. 
    \item Our method effectively \textbf{prevents outliers} that is much more likely to occur for previous approaches.
    \item Our method \textbf{accelerates optimization} and reduces the inference time of the previous gradient-based approach.
    \item Experiments on synthetic and real-world datasets demonstrate the \textbf{superior performance} of our method with a variety of point clouds and noise types. 
\end{enumerate}

\section{related work}
\label{sec:related}

To perform point cloud denoising, traditional methods~\cite{alexa2001point,avron2010,cazals2005estimating, huang2013edge,mattei2017point,sun2015denoising} heavily rely on geometric priors but show limited performance. Deep learning-based approaches break the performance limit of the point cloud denoising. Among them, some~\cite{duan20193d, hermosilla2019total,pistilli2020learning,rakotosaona2020pointcleannet,li2023single} denoise by estimating the deviation of noisy points from the clean surface, but often results in outliers due to the coarse one-step estimation. Others~\cite{luo2020differentiable} predict the underlying manifold of a noisy point cloud for reconstruction, which loses details in the downsampling stage. 

Recently, Score~\cite{luo2021score} is proposed to tackle the aforementioned issues by iteratively updating the point position in implicit gradient fields learned by neural networks. 
% Given a noisy point cloud $\Xmat = \{ \xv _i\}_{i=1}^N$ consisting of $N$ points as input, Score assumes the underlying clean point cloud is sampled from a distribution $p$, and the noise follows a distribution $n$. Then Score estimates the gradient of the log-probabilityfunction, \textit{i.e.}, $\bigtriangledown_{\xv}\log[(p * n)(\xv)]$, from $\Xmat$ using a neural network, and then denoise $\Xmat$ momentum gradient ascent, thus moving noisy points towards the mode of the distribution that corresponds to the underlying clean surface.
However, the predicted gradients suffer fluctuation, leading to perturbed and unstable solutions, as well as a large inference time. Moreover, once its estimated gradient deviates from the correct direction, the gradient estimation errors could be accumulated and result in serious outliers. 

\begin{figure*}
\centering
\vspace{-15pt}
    \includegraphics[width =0.9\linewidth]{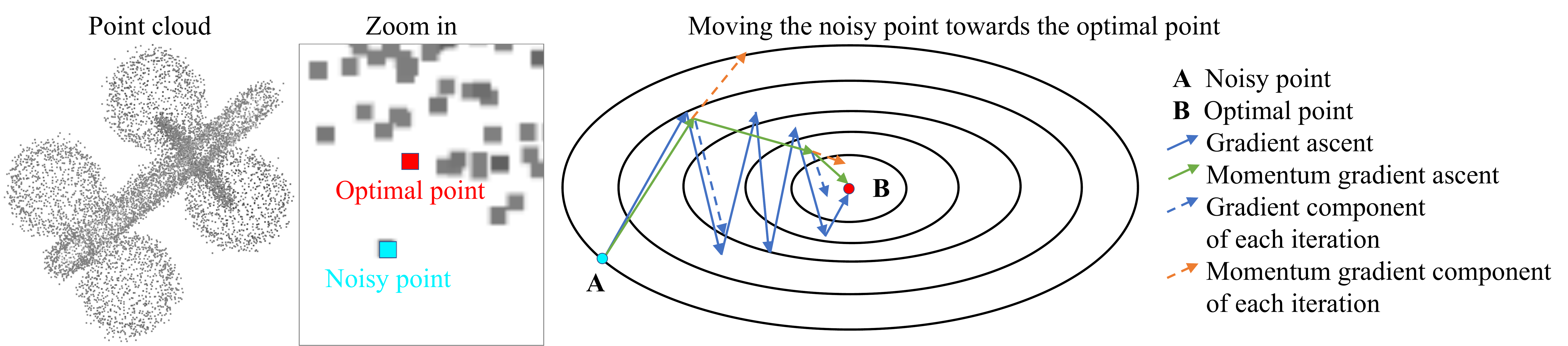}
    \vspace{-5pt}
    \caption{Comparison of classical gradient ascent and momentum gradient ascent.}
    \label{fig:ascent}
\end{figure*}

\begin{table*}[]
	\centering
 \vspace{-6pt}
	\resizebox{0.88\textwidth}{!}{
\begin{tabular}{c|c|cccccc|cccccc}
\midrule
\multicolumn{2}{c}{\# Points} & \multicolumn{6}{c}{10K (Sparse)} & \multicolumn{6}{c}{50K (Dense)} \\
% \cline{2-13}
\multicolumn{2}{c}{Noise} & \multicolumn{2}{c}{1\%} &
\multicolumn{2}{c}{2\%} & \multicolumn{2}{c}{3\%} & \multicolumn{2}{c}{1\%} &
\multicolumn{2}{c}{2\%} & \multicolumn{2}{c}{3\%}\\
Type & Method & CD & P2M & CD & P2M & CD & P2M & CD & P2M & CD & P2M & CD & P2M \\
% \midrule
% Noise Type & \multicolumn{12}{c}{Gaussian}\\
\midrule
\multirow{6}{*}{Gaussian}&
Bilateral~\cite{digne2017bilateral} & 3.646 & 1.342 & 5.007 & 2.018 & 6.998 & 3.557 & 0.877 & 0.234 & 2.376 & 1.389 & 6.304 & 4.730 \\
&GLR~\cite{zeng20193d} &2.959&1.052&3.773&1.306&4.909&2.114&0.696&0.161&1.587&0.830&3.839&2.707\\
&PCNet~\cite{rakotosaona2020pointcleannet} & 3.515 & 1.148 & 7.467 & 3.965 & 13.067 & 8.737 & 1.049 & 0.346 & 1.447 & 0.608 & 2.289 & 1.285  \\
&DMR~\cite{luo2020differentiable} &4.482&1.722&4.982&2.115&5.892&2.846&1.162&0.469&1.566&0.800&2.432&1.528  \\
&Score~\cite{luo2021score} & 2.521&0.463&3.686&1.074&4.708&1.942&0.716&0.150&1.288&0.566& \B{1.928} & \B{1.041} \\
&Ours & \B{2.498} & \B{0.459} & \B{3.629} & \B{1.054} & \B{4.686} & \B{1.923} & \B{0.706} & \B{0.146} & \B{1.287} & \B{0.557} & 1.931 & 1.045 \\
% \midrule
% Noise Type & \multicolumn{12}{c}{Laplace}\\
\midrule
\multirow{5}{*}{Laplace}&
GLR~\cite{zeng20193d} &3.223&1.121&4.751&2.090&7.977&4.773&0.962&0.374&3.269&2.325&8.675&7.162 \\
&PCNet~\cite{rakotosaona2020pointcleannet} & 4.616&1.940&11.082&7.218&20.981&15.922& 1.190&0.458&2.854&1.868&7.555&6.020  \\
&DMR~\cite{luo2020differentiable} &4.600&1.811&5.441&2.469&6.918&3.714&1.243&0.537&1.881&1.077&3.609&2.634  \\
&Score~\cite{luo2021score} & 2.915&0.674&4.601&1.799&6.332&3.271& 0.823&0.231&1.658&0.869&2.728&1.681 \\
&Ours & \B{2.887} & \B{0.669} & \B{4.521} & \B{1.754} & \B{6.237} & \B{3.257}& \B{0.816} & \B{0.212} & \B{1.648} & \B{0.857} & \B{2.717} & \B{1.679} \\
% \midrule
% Noise Type & \multicolumn{12}{c}{Uniform}\\
\midrule
\multirow{5}{*}{Uniform}&
GLR~\cite{zeng20193d} &1.850&1.015&2.948&1.052&3.400&1.109&\B{0.485}&0.071&\B{0.656}&0.132&\B{0.903}&0.293 \\
&PCNet~\cite{rakotosaona2020pointcleannet} & \B{1.205}&0.337&3.378&1.018&5.044&1.995& 0.806&0.228&1.064&0.358&1.218&0.451  \\
&DMR~\cite{luo2020differentiable} &4.307&1.640&4.445&1.693&4.685&1.857&1.064&0.391&1.159&0.464&1.287&0.572  \\
&Score~\cite{luo2021score} & 1.277&0.248&2.467&0.418&3.079&0.654& 0.506&0.047&0.690&0.129&0.917&0.282 \\
&Ours & 1.243&\B{0.239}&\B{2.404}&\B{0.397}&\B{3.052}&\B{0.638} & 0.487&\B{0.045}&0.679&\B{0.122}&0.905&\B{0.278}\\
 \bottomrule
\end{tabular}
}
\vspace{-5pt}
	\caption{Point cloud denoising comparisons on the PU-Net dataset~\cite{yu2018pu}. CD and P2M are multiplied by $10^4$.
	}
\label{table:exp}
\end{table*}

\section{method}
\label{sec:method}

Given a noisy point cloud $\Xmat = \{ \xv \}_{i=1}^N$, 
we first implement a network that inputs noisy point clouds and outputs point-wise gradients, and then utilize the estimated gradients to denoise point clouds by momentum ascent, as Figure~\ref{fig:method} shows.

\subsection{Neural Network and Loss Function}
\label{sec:network}

 The network aims at estimating the gradient in the neighborhood space around $\xv_i$, denoted as $\hat{g}_i(\xv)$. We adopt the network and loss function reported in Score~\cite{luo2021score}, which inputs point coordinates $\xv \in \mathbb{R}^3$ surrounding $\xv_i$, learns point-wise features $f_i$, and outputs the estimated gradient $\hat{g}_i(\xv)$:
\begin{align}
    \hat{g}_i(\xv) = G(\xv - \xv_i, f_i),
\end{align}
where $G(\cdot)$ is the gradient estimation network implemented by a multi-layer perceptron.
% Following Score~\cite{luo2021score}, we denote the ground truth clean point cloud as $\Ymat =
% \{ \yv_i\}^N_{i=1}$, and define the gradient for point $\xv$ as
% \begin{align}
%     g(\xv) = N(\xv, \Ymat) - \xv, \qquad \xv \in \mathbb{R}^3,
% \end{align}
% where $N(\xv, \Ymat)$ is the point nearest to $\xv$ in $\Ymat$, $g(x)$ is a vector from $\xv$ to the clean surface. 
Then the loss function~\cite{luo2021score} is
\begin{align}
    \mathcal{L} = \frac{1}{N} \sum^{N}_{i=1}  \mathcal{L}_i = \frac{1}{N} \mathbb{E}_{\xv \sim \mathcal{N}(\xv_i)}
    \big[\ || g(\xv) - \hat{g}_i(\xv) ||_2^2\ \big], 
\end{align}
where $\mathcal{N}(\xv_i)$ is a distribution concentrated in the neighborhood of $\xv_i$ in $\mathbb{R}^3$ space, $g(x)$ is a vector from $\xv$ to the ground truth clean surface. 
Following Score~\cite{luo2021score}, the ensemble gradient is calculated as
\begin{align}
\begin{aligned}
        \zv_i(\xv) = \frac{1}{k} \sum_{\xv_j} \hat{g}_j(\xv), \qquad \xv_j \in H(\xv_i; k),\ \xv \in \mathbb{R}^3,
\end{aligned}
\end{align}
where $H(\xv_i; k)$ is the k-nearest neighborhood of $\xv_i$.

Though Score~\cite{luo2021score} uses the ensemble gradient for robustness, it still suffers in fluctuated gradients. Moreover, once its estimated gradient deviates from the correct direction, the errors could be accumulated and result in serious outliers. 

\subsection{Momentum Ascent Denoising Algorithm}
\label{sec:ascent}

To alleviate those problems faced by previous gradient-based methods, we propose to perform point cloud denoising by updating point coordinates with momentum gradient ascent.
At the beginning of point cloud denoising, we initialize the coordinate for each point according to the input point cloud:
\begin{align}
    \begin{aligned}
        \xv_i^{(0)} &= \xv_i, \qquad \xv_i \in \Xmat.
    \end{aligned}
\end{align}

To leverage past gradients, we introduce an auxiliary vector, which is initialized as zero and updated with a leaky average over past gradients:
\begin{align}
    \begin{aligned}
       \vv_i^{(0)}& = \mathbf{0} \in \mathbb{R}^3, \\
       \vv_i^{(t)} &= \alpha \zv_i(\xv_i^{(t-1)}) + (1-\alpha) \vv_i^{(t-1)}, \ t = 1, \dots, T,
    \end{aligned}
\label{eq:v}
\end{align}
where $t$ is the iterative step, $ \alpha$ is the momentum weight, $\vv_i$ serves to relieve gradient variance and obtain more stable directions of ascent.
Finally, the point cloud is denoised by updating point coordinates with momentum gradient ascent:
\begin{align}
    \begin{aligned}
        \xv_i^{(t)} &= \xv_i^{(t-1)} + \beta \gamma^t \vv_i^{(t)},
        \qquad t = 1, \dots, T,
    \end{aligned}
\label{eq:x}
\end{align}
where $t$ is the iterative step, $T$ is the total number of iteration steps, $\beta$ is the step size, $\gamma^t$ is the decay coefficient decreasing towards $0$ to ensure convergence. While Score requires a relatively large number of total steps $T = 30$ to conduct their experiments, our method applies momentum gradient ascent, thus reducing the step number to $T=15$ while achieving better performance. 

As Figure~\ref{fig:ascent} shows, compared to classical gradient ascent, momentum gradient ascent leverages the information of previous iterations in determining the trajectories of points, thereby benefiting solution stability and inference time.

\begin{figure}
\centering
\vspace{-5pt}
    \includegraphics[width =\linewidth]{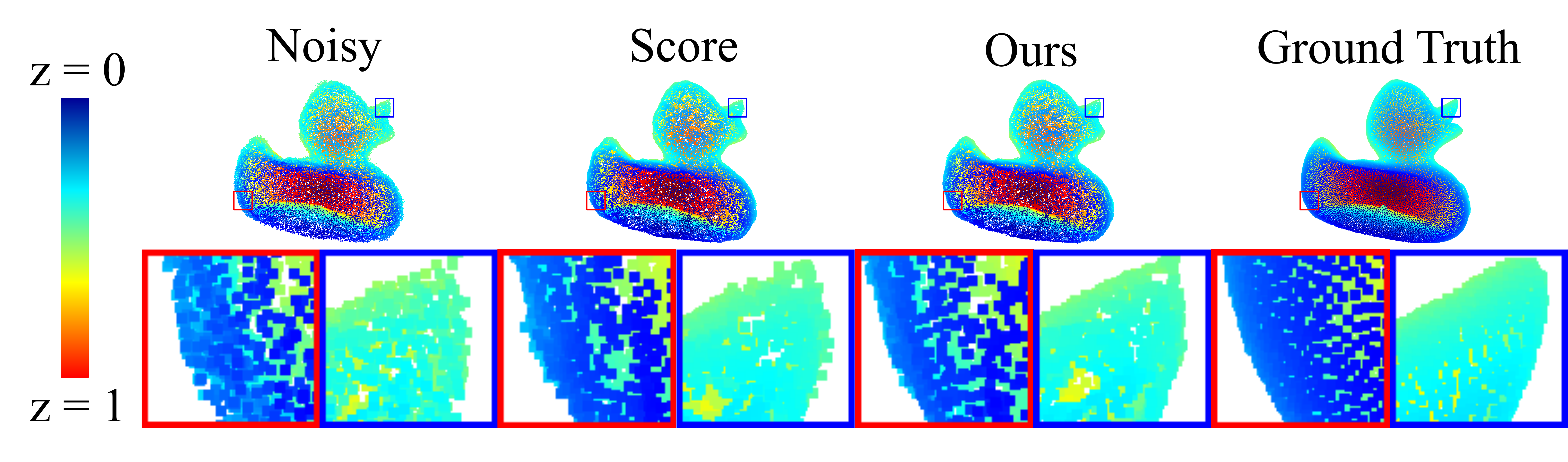}
    \vspace{-18pt}
    \caption{Comparisons on the dense point cloud \texttt{duck} of the PU-Net dataset~\cite{yu2018pu}. The color of each 3D point varies according to its $z$ coordinate (perpendicular to the plane).}
    \label{fig:duck}
\end{figure}

\begin{figure*}
\centering
\vspace{-15pt}
    \includegraphics[width =\linewidth]{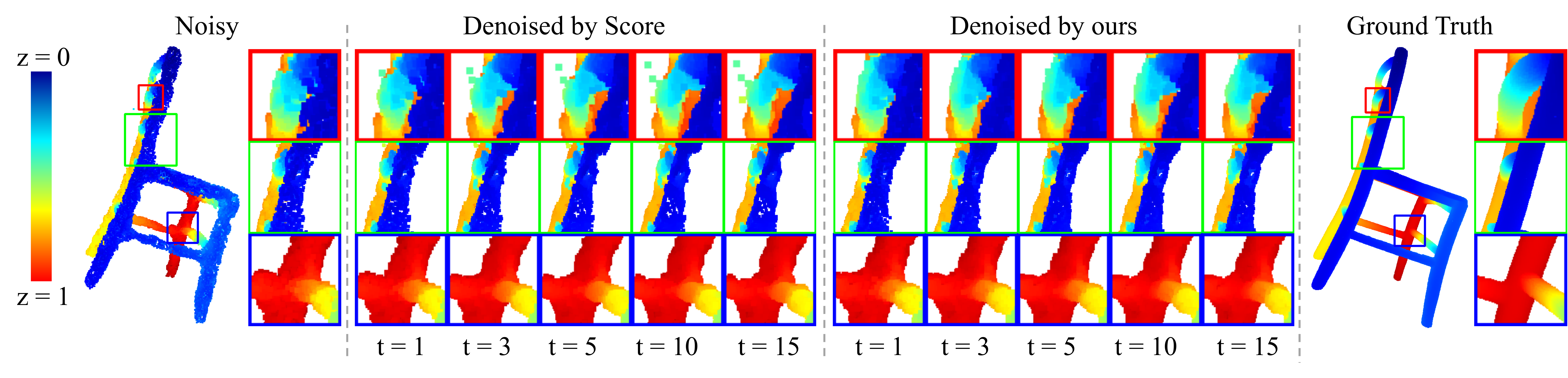}
    \vspace{-20pt}
    \caption{Point cloud denoising comparisons on the dense point cloud \texttt{chair} of the PU-Net dataset~\cite{yu2018pu}. The color of each 3D point varies according to its $z$ coordinate (perpendicular to the plane). }
    \label{fig:ascent_exp}
\end{figure*}

\begin{figure*}[]
\centering
\vspace{-5pt}
    \includegraphics[width =\linewidth]{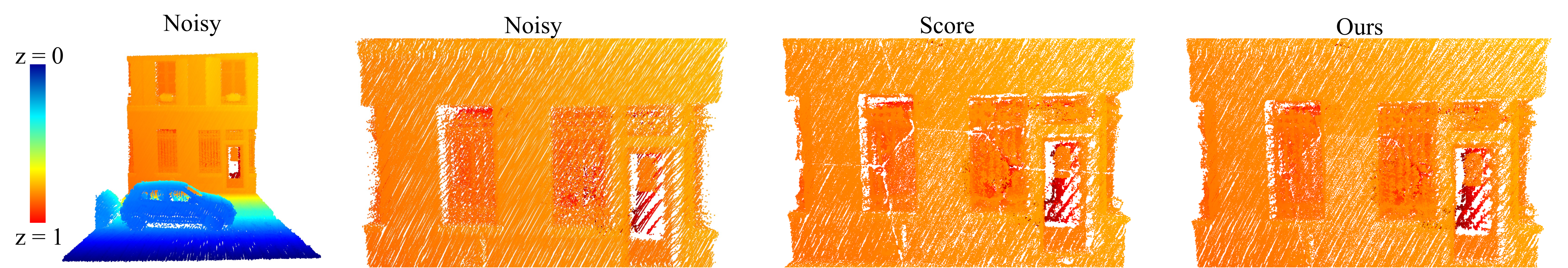}
    \vspace{-15pt}
    \caption{Point cloud denoising comparisons on the real-world dataset \texttt{Paris-rue-Madame}~\cite{yu2018pu}. The color of each 3D point varies according to its $z$ coordinate (perpendicular to the plane).}
    \label{fig:real_data}
    \vspace{-5pt}
\end{figure*}

\section{experiment}
\label{sec:exp}

% \subsection{Experimental Setting}
\noindent{\textbf{Experiment Setting.}} 
The commonly used PU-Net dataset~\cite{yu2018pu} is adopted for training and testing following previous works~\cite{rakotosaona2020pointcleannet,luo2021score,luo2020differentiable}. Moreover, we use the real-world dataset Paris-rue-Madame~\cite{serna2014paris} for qualitative evaluation. 
We utilized a single model trained on the PU-Net training set to conduct both synthetic and real-world experiments. 
% In the case of real-world one, ground truths are not provided for supervision, making it a challenging task that verifies the generalization and robustness of denoising methods.
Two traditional methods, bilateral filtering~\cite{digne2017bilateral}, GLR~\cite{zeng20193d}, and three deep-learning-based approaches including PCNet~\cite{rakotosaona2020pointcleannet}, DMR~\cite{luo2020differentiable} and Score~\cite{luo2021score} are used for comparison. 
Two commonly used metrics are adopted for quantitative evaluation, the Chamfer distance (CD)~\cite{fan2017point} and point-to-mesh distance (P2M)~\cite{ravi2020accelerating}.
% We use the PU-Net dataset~\cite{yu2018pu} for training and testing, following previous works~\cite{rakotosaona2020pointcleannet,luo2021score,luo2020differentiable}, and employ the real-world dataset Paris-rue-Madame~\cite{serna2014paris} for qualitative evaluation. 
% A single model trained on the PU-Net dataset is utilized for both synthetic and real-world experiments. The real-world task provides no ground truth for supervision, making it a challenging test that verified the generalization and robustness of denoising methods.
% We compar our approach against two traditional methods, bilateral filtering~\cite{digne2017bilateral} and GLR~\cite{zeng20193d}, as well as three deep-learning-based approaches: PCNet~\cite{rakotosaona2020pointcleannet}, DMR~\cite{luo2020differentiable}, and Score~\cite{luo2021score}.
% We employ two commonly used metrics, the Chamfer distance (CD)~\cite{fan2017point} and point-to-mesh distance (P2M)~\cite{ravi2020accelerating}, for quantitative evaluation.
\newline
% \subsection{Quantitative and Qualitative Results}
\noindent{\textbf{Quantitative and Qualitative Results.}}
As Table~\ref{table:exp} shows, our model significantly outperforms previous methods bilateral filtering~\cite{digne2017bilateral}, PCNet~\cite{rakotosaona2020pointcleannet}, DMR~\cite{luo2020differentiable}, and surpasses GLR~\cite{zeng20193d}, Score~\cite{luo2021score} in the majority of cases. 
To further illustrate the effectiveness of our method, we compare the qualitative performance with the competitive baseline Score~\cite{luo2021score}. 
As Figure~\ref{fig:duck} shows, Score takes 30 steps to achieve the denoising results and suffers in outliers, while ours takes only 15 steps and outputs results with fewer outliers and smoother outlines.
In Figure~\ref{fig:ascent_exp}, compared to Score~\cite{luo2021score}, our denoised point clouds feature fewer outliers, better structures and smoother outlines in all the iterative steps. 
In Figure~\ref{fig:real_data}, Score results in cracks, while ours effectively reduces the outliers, maintains cleaner and smoother outlines.
% \newline

\begin{table}[]
	\centering
	\resizebox{0.3\textwidth}{!}{
\begin{tabular}{c|c|c}
\midrule
\# Points & 10K (Sparse) & 50K (Dense) \\
% \# Iter & 1 & 2 & 1  & 2 \\
\midrule
% Score & 39.73 & 3760.21\\
% Ours & 24.23 & 2855.37 \\
Score~\cite{luo2021score} & 0.66 & 6.27\\
Ours & \B{0.40} & \B{4.76} \\
 \bottomrule
\end{tabular}
}
\vspace{-5pt}
\caption{Average inference time per point cloud in minutes on PU-Net~\cite{yu2018pu} dataset using identical environments.
	}

\label{table:runtime}
\end{table}

\begin{table}[]
	\centering
	\tabcolsep=0.2cm
	\resizebox{0.45\textwidth}{!}{
\begin{tabular}{c|c|c|c|c|c|c|c|c}
\midrule
% \cline{2-13}
\multicolumn{3}{c}{Gaussian Noise} & \multicolumn{2}{c}{1\%} &
\multicolumn{2}{c}{2\%} & \multicolumn{2}{c}{3\%} \\
$T$ & $\alpha$ & $\beta$ & CD & P2M & CD & P2M & CD & P2M \\
\midrule
$15$ & $0.9$ & $0.2$ & 2.498 & 0.459 & \B{3.629} & \B{1.054} & \B{4.686} & \B{1.923}\\%%%
\midrule
$1$ & $0.9$ & $0.2$ &2.858&0.695&5.665&2.576&7.905&4.467\\%%%

$5$ & $0.9$ & $0.2$ &2.489&0.453&3.998&1.283&4.906&2.038\\%%%

$10$ & $0.9$ & $0.2$ &2.488&0.453&3.784&1.137&4.725&1.938\\%%%

$20$ & $0.9$ & $0.2$ &2.508&0.463&3.702&1.091&4.696&1.940\\%%%

$30$ & $0.9$ & $0.2$ &2.522&0.468&3.683&1.084&4.705&1.946\\%%%
\midrule
$15$ & $0.5$ & $0.2$ &2.500&0.463&3.732&1.109&4.724&1.959\\%%%

$15$ & $0.8$ & $0.2$ &2.499&0.460&3.728&1.104&4.702&1.932\\%%%

$15$ & $1.0$ & $0.2$ &2.521&0.463&3.686&1.074&4.708&1.942\\%%%
\midrule
$15$ & $0.9$ & $0.1$ &\B{2.482}& \B{0.446} &3.948&1.244&4.838&1.984\\%%%

$15$ & $0.9$ & $0.3$ &2.526&0.470&3.680&1.083&4.704&1.946\\

$15$ & $0.9$ & $0.5$ &2.594&0.499&3.700&1.116&4.864&2.087\\
 \bottomrule
\end{tabular}
}
\vspace{-5pt}
	\caption{Ablation study \textit{w.r.t.} hyper-params $T, \alpha, \beta$ on sparse point clouds from PU-Net~\cite{yu2018pu}.  CD \& P2M multiplied $10^4$. 
	}
\label{table:as}
% \vspace{-15pt}
\end{table}
% \subsection{Inference Time}
\noindent{\textbf{Inference Time.}}
As Table~\ref{table:runtime} shows, Score requires a longer inference time. In contrast, our method utilizes previous gradients to seek promising directions to move forward, thus reducing inference time by approximately 25\% $\sim$ 40\%.
\newline
% \subsection{Ablation Study}
\noindent{\textbf{Ablation Study.}}
We investigate hyper-parameters of the proposed algorithm formulated in Equation~\ref{eq:v}, \ref{eq:x}. Other implementation details, including learning rates, decay coefficient $\gamma$, \textit{etc.}, are the same as the classical gradient-based method Score~\cite{luo2021score}.
According to Table~\ref{table:as}, we recommend the setting $T=15$, $\alpha=0.9$, and $\beta=0.2$, which is used in this paper.

\section{Conclusion}
In this paper, we propose point cloud denoising via momentum ascent in gradient fields. To improve the previous gradient-based method, we propose a simple yet effective iterative paradigm of point cloud denoising, which leverages past gradients to seek promising directions to move forward. Our method effectively prevents outliers that are much more likely to occur for previous methods. Moreover, our method accelerates optimization and reduces the inference time of the previous gradient-based method. On both synthetic and real-world datasets, extensive experiments demonstrate that our method outperforms state-of-the-art methods with a variety of point clouds, noise types and noise levels.

\bibliographystyle{IEEEbib}
\bibliography{refs}

\end{document}